# Fast Multi-class Dictionaries Learning with Geometrical Directions in MRI Reconstruction

Zhifang Zhan, Jian-Feng Cai, Di Guo, Yunsong Liu, Zhong Chen, Xiaobo Qu*

*Abstract*— **Objective:** Improve the reconstructed image with fast and multi-class dictionaries learning when magnetic resonance imaging is accelerated by undersampling the k-space data. *Methods:* A fast orthogonal dictionary learning method is introduced into magnetic resonance image reconstruction to providing adaptive sparse representation of images. To enhance the sparsity, image is divided into classified patches according to the same geometrical direction and dictionary is trained within each class. A new sparse reconstruction model with the multi-class dictionaries is proposed and solved using a fast alternating direction method of multipliers. *Results:* Experiments on phantom and brain imaging data with acceleration factor up to 10 and various undersampling patterns are conducted. The proposed method is compared with state-of-the-art magnetic resonance image reconstruction methods. *Conclusion:* Artifacts are better suppressed and image edges are better preserved than the compared methods. Besides, the computation of the proposed approach is much faster than the typical K-SVD dictionary learning method in magnetic resonance image reconstruction. *Significance:* The proposed method can be exploited in undersapmled magnetic resonance imaging to reduce data acquisition time and reconstruct images with better image quality.

## I. INTRODUCTION

THE compressed sensing (CS) theory proved that a sparse signal can be accurately reconstructed from a small number of random measurements [1, 2]. In magnetic resonance imaging (MRI), imaging speed is critical for applications. Thus, CS is introduced into MRI and has significantly reduced the data acquisition time [3]. This new imaging technology is called CS-MRI for short. Its combination with other fast MRI methods, e.g. parallel imaging [4-9], non-Cartesian sampling [10-12], low rank [13-15] and non-convex optimization [16-19], can further speed up imaging.

Finding an optimal sparse representation of images is an active research area in CS-MRI since a sparser representation usually leads to lower reconstruction error [20, 21]. Pre-specified dictionaries usually capture only one type of image features, and reconstruction qualities are not satisfactory. For example, contourlets [22] and bandelets [23] are applicable to piecewise smooth images with smooth boundary and/or geometrical directions. Combination of wavelets, contourlets and total variation [24] can suppress the artifacts produced by one transform but there is still loss of image structures when data are highly undersampled.

Recently, adaptive dictionaries or transforms are explored by enforcing the sparsity on image patches [20, 21, 25-28], which has significantly improved the reconstructed image quality than that using pre-specified dictionaries. K-SVD [29] is a typical dictionary learning method which has been applied in CS-MRI for a single image [20, 27, 30] or image series [31-34]. However, these methods are time consuming in the iterative magnetic resonance (MR) image reconstructions [20, 29] and may fail to sparsely represent some patches that are excluded in dictionary training. Fortunately, the computation of K-SVD can be efficiently reduced by accelerating the sparse coding step [35] or with quicker approximation of the exact singular value decomposition [36]. These modified K-SVD methods however have not been investigated in MRI so far, thus their performances are still unknown.

In this paper, we propose a Fast Dictionary Learning method on Classified Patches (FDLCP) to reconstruct MR image from highly undersampled data. The dictionaries training is implemented by a small-scale singular value decomposition (SVD) and a thresholding operation, making it computationally efficient. To improve the sparsity, multi-class dictionaries are trained on the classified image patches according to their geometrical directions. A sparse image reconstruction model is proposed on the multi-class dictionaries in CS-MRI. Overall, the proposed method makes use of both the similarity and the geometrical directions of patches and provides a sparser approximation for the target image.

To illustrate the benefits of the proposed method, we carry out experiments on both phantom and brain MRI data. The experiments show that the proposed classified dictionaries provide a sparser representation than non-classified adaptive dictionary. Moreover, it outperforms state-of-the-art MR reconstruction methods including dictionary learning MRI (DLMRI) [20], wavelet tree sparsity MRI (WaTMRI) [37], and patch-based directional wavelets (PBDW) [21], in reducing artifacts, minimizing reconstruction error and saving computational time.

This work is supported by the NNSF of China (61571380, 61201045, 61302174 and 11375147), Natural Science Foundation of Fujian Province of China (2015J01346), Fundamental Research Funds for the Central Universities (20720150109, 2013SH002), Important Joint Research Project on Major Diseases of Xiamen City (3502Z20149032), and NSF DMS-1418737. (*Corresponding author: Xiaobo Qu*)

Zhifang Zhan, Yunsong Liu, Zhong Chen and Xiaobo Qu are with the Department of Electronic Science, Fujian Provincial Key Laboratory of Plasma and Magnetic Resonance, Xiamen University, Xiamen, China (e-mail: zhanzfadu@stu.xmu.edu.cn;yunsongliu@stu.xmu.edu.cn;chenz@xmu.edu.cn; quxiaobo@xmu.edu.cn)

Jian-Feng Cai is with Department of Mathematics, Hong Kong University of Science and Technology, Hong Kong SAR, China. This work was partially done when J. F. Cai was at Department of Mathematics, University of Iowa, Iowa City, USA. (e-mail: jfcai@ust.hk)

Di Guo is with the School of Computer and Information Engineering, Fujian Provincial University Key Laboratory of Internet of Things Application Technology, Xiamen University of Technology, Xiamen, China (e-mail: guodi@xmut.edu.cn)

The rest of the paper is organized as follows. In Section II, we briefly review the CS-MRI technology and fast dictionary training algorithms. We propose the multi-class dictionaries sparse reconstruction model for CS-MRI and derive an efficient iterative algorithm in Section III. Section IV demonstrates the performance of the proposed method. Finally, we make the conclusion and discuss the future work in Section V.

## II. BACKGROUND AND RELATED WORK

### A. CS-MRI

Let $\mathbf{x} \in \mathbb{C}^{N^2}$ be a $N \times N$ MR image in a vector form to be reconstructed, $\mathbf{F_U} \in \mathbb{C}^{M \times N^2} \left( M < N^2 \right)$ be the undersampled Fourier encoding matrix, and $\mathbf{y} = \mathbf{F_U} \mathbf{x} \in \mathbb{C}^M$ represents the undersampled k-space data. MR images can be reconstructed from undersampled data by employing the sparse reconstruction model. Define $\mathbf{\Psi} \in \mathbb{C}^{T \times N^2}$ as the sparsifying transform, where images have sparse representations under this transform. A typical CS-MRI reconstruction is obtained by solving the following problem [3]:

$$\min_{\mathbf{x}} \|\mathbf{\Psi x}\|_1 \quad s.t. \quad \|\mathbf{y} - \mathbf{F_U x}\|_2 \le \varepsilon, \quad (1)$$

where $\varepsilon$ is a parameter controlling the fidelity of the reconstruction to the measured data. Minimizing the $l_1$ norm $\|\mathbf{\Psi x}\|_1$ promotes the image sparse representation and the $l_2$ norm constraint $\|\mathbf{y} - \mathbf{F_U x}\|_2 \le \varepsilon$ enforces the data consistency. Equation (1) tries to find the sparsest representation among all possible solutions that are consistent with the acquired data.

### B. Fast Dictionary Learning (FDL)

In most of dictionary learning methods, adaptive dictionaries are trained from image patches [29, 38]. The basic idea of these approaches is to train a set of atoms, columns in the dictionary, from image patches so that these patches can be approximated by a sparse linear combination of these atoms.

Let $\mathbf{D} \in \mathbb{C}^{n^2 \times k}$ denote the adaptive dictionary, $\boldsymbol{\alpha}_i \in \mathbb{C}^k$ is the sparse representation of an image patch $\mathbf{x}_i$ with respect to dictionary $\mathbf{D}$. The popular K-SVD method [29] trains an adaptive dictionary by solving the following minimization problem

$$\min_{\mathbf{D}, \{\boldsymbol{\alpha}_i\}} \sum_{i=1}^{q} \|\mathbf{x}_i - \mathbf{D}\boldsymbol{\alpha}_i\|_2^2 \quad s.t. \quad \|\boldsymbol{\alpha}_i\|_0 \le P_0, \forall i \in \{1, 2, \cdots, q\} \quad (2)$$

where $q$ is the number of the trained image patches, $P_0$ is a given sparsity level and the columns of $\mathbf{D}$, also called atoms, are constrained to have unit norm to avoid the scaling ambiguity [20, 29, 41]. The K-SVD alternates between sparse coding of the examples based on the current dictionary and updating the dictionary atoms to fit the data. An overcomplete dictionary, meaning $k > n^2$, is commonly trained although K-SVD is not restricted to this. The original K-SVD method has been adopted in CS-MRI [20]. And this adaptive reconstruction framework has shown superior performance than the non-adaptive reconstruction [20]. However, one problem of the original K-SVD in CS-MRI is its relatively low training speed [39].

The dictionary training procedure can be accelerated with smarter algorithms. For examples, multiple atoms and sparse coefficients are simultaneously updated in [40, 41] while the majorization method [42] and the first order series expansion for factorization [43] are also incorporated to speed up the training process. These dictionary training methods however have not been investigated in CS-MRI so far, thus their reconstruction performances are still unknown.

Reducing the dictionary dimension provides another way of fast learning, meaning $k \le n^2$. Recently, orthogonal dictionaries $\mathbf{D}$ [38, 44], satisfying $\mathbf{D}^H \mathbf{D} = \mathbf{I}$, or unitary dictionary learning [45, 46], are explored for image denoising. Nearly orthogonal dictionaries or transforms learning are also developed in [47-49]. It has been shown that these methods can achieve comparable or even better performance in image denoising than the original K-SVD but runs much faster [38, 44-49]. Due to nice property of orthogonality, the orthogonal dictionary is expected to enable both fast computation and adaptive sparse representation in CS-MRI.

Let $\mathbf{X} = \left[ \mathbf{x}_1, \mathbf{x}_2, \cdots, \mathbf{x}_q \right] \in \mathbb{C}^{n^2 \times q}$ denote the set of trained images patches, $\mathbf{A} = \left[ \boldsymbol{\alpha}_1, \boldsymbol{\alpha}_2, \cdots, \boldsymbol{\alpha}_q \right] \in \mathbb{C}^{n^2 \times q}$ be the sparse approximation of images patches $\mathbf{X}$ under the orthogonal dictionary $\mathbf{D}$. The orthogonal dictionary is learnt [38] as

$$\min_{\mathbf{D}, \mathbf{A}} \|\mathbf{X} - \mathbf{DA}\|_F^2 + \eta^2 \|\mathbf{A}\|_0 \quad s.t. \quad \mathbf{D}^H \mathbf{D} = \mathbf{I}. \quad (3)$$

where $\|\cdot\|_F$ is the Frobenius norm of a matrix, $\|\mathbf{A}\|_0$ denotes the number of nonzero entries in $\mathbf{A}$. Equation (3) is solved by alternatively computing the sparse coding $\mathbf{A}$ with simple hard thresholding and updating the dictionary $\mathbf{D}$ with a SVD decomposition. Hence the dictionary learning algorithm is simple and the whole training process is much faster than the commonly used K-SVD [38]. We refer to this dictionary learning method [38, 44] as Fast Dictionary Learning (FDL). Therefore, using FDL method in CS-MRI is supposed to consume less computation time than that using the original K-SVD dictionary learning method in CS-MRI [20].

In this paper, the fast dictionary learning will be introduced into MR image reconstruction. To enhance the sparsity, image is divided into classified patches according to the same geometrical direction and dictionary is trained within each class. We set up a sparse reconstruction model with the multi-class dictionaries and solve the problem with a fast alternating direction method of multipliers. It is worth noting that nearly unitary dictionary or transforms learning have been applied to CS-MRI [39, 50] that run much faster than the typical dictionary learning MRI [20]. We tried to compare the results with [39, 50] but the codes are not available from authors. However, our proposed method differs greatly from [39, 50] as follows: 1) We learn multi-class dictionaries on classified image patches, rather than a single dictionary/transform; 2) We learn dictionaries from an approximately reconstructed image before the iterative MR image reconstruction, instead of training the dictionary and reconstructing the image simultaneously in the iterative reconstruction; The promising performance of the proposed method is comprehensively compared with other state-of-the-art MR reconstruction methods in CS-MRI problems.

### III. PROPOSED METHOD

#### A. Fast Dictionary Learning on Classified Patches

Two properties of adaptive dictionary are expected. First, it is able to enforce patches sparsity of the target image. Second, the learning process is computationally efficient. The latter has been solved by SVD with hard thresholding [38], but how to provide an optimal sparse representation is still challenging. Since image patches contain substantial and distinct features, an adaptive dictionary learnt from all the images patches may not capture all the valid image features sufficiently. On the other hand, multi-class dictionaries learning has shown ability to sparsely represent distinct image features [49, 51], where image patches are classified by containing the incoherence on inter-class dictionaries [51] or minimizing the sparse approximation error with the optimal class dictionary [49]. Both methods [49, 51] classify patches in the process of dictionaries learning and have not been investigated in the CS-MRI problem.

In this paper, we propose a distinctive way of multi-class dictionary learning and reconstructing MR images from undersampled data. Image patches are classified using explicit geometrical directions estimated from a pre-reconstructed image, and then each class of orthogonal dictionary is learnt on patches within each class. These multi-class dictionaries are fixed in the iterative MR image reconstruction, saving a lot of computation on dictionary learning.

The proposed method incorporates patches information to benefit dictionary learning, inspired by estimating geometrical directions in sparse representation of images [21, 23]. We choose the geometrical directions estimation proposed in [23] because the computation is fast while preserving the image directions very well for CS-MRI [21].

An optimal direction $\omega_j$ of $j^{th}$ patch is estimated with

$$\min_{\omega \in \{\theta_1, \cdots, \theta_Q\}} \left\| \hat{\mathbf{c}}_{j,\omega} - \mathbf{W}^T \mathbf{G}_\omega \mathbf{x}_j \right\|_2^2 \quad (4)$$

where $\mathbf{G}_\omega$ is an operator that re-arrange pixels along a candidate geometrical direction $\{\theta_1, \theta_2, \cdots, \theta_Q\}$ to form a 1D vector [21] (The detailed description of $\mathbf{G}_\omega$ is in Appendix A), $\mathbf{W}^T$ is the forward 1D Haar wavelets to provide the sparse representation of these re-arranged pixels that are in the form of 1D vectors, and $\hat{\mathbf{c}}_{j,\omega}$ is the preserved 25% largest wavelet coefficients. The geometrical direction provide optimal sparsity among candidate directions. Details of the directions estimation can be found in [21, 23]. As shown in Fig. 1(a) and (b), the red lines are the estimated directions of image patches.

Dictionary training is performed on patches of the same class which are classified according to their geometrical directions. For example, one class of patches with diagonal geometrical direction is formed in Fig. 1(c). An orthogonal dictionary is trained in each class according to

$$\min_{\mathbf{D}_\omega, \mathbf{A}} \left\| \mathbf{X}_\omega - \mathbf{D}_\omega \mathbf{A} \right\|_F^2 + \eta^2 \left\| \mathbf{A} \right\|_0 \quad s.t. \quad \mathbf{D}_\omega^H \mathbf{D}_\omega = \mathbf{I} \quad (5)$$

where $\mathbf{D}_\omega$ is the dictionary for the patches $\mathbf{X}_\omega$ that shares the same geometrical direction $\omega$. Equation (5) is solved by alternatively computing the sparse coding $\mathbf{A}$ and updating the dictionary $\mathbf{D}_\omega$ with SVD in each iteration [38, 45, 46, 52].

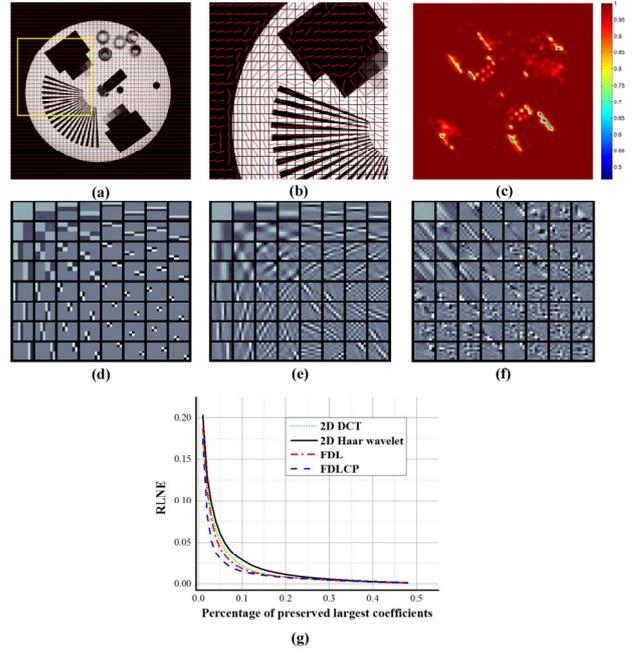

Fig. 1. Comparison on the sparsity using 2D Haar wavelets, 2D DCT, adaptive fast dictionary learning (FDL) and adaptive fast dictionary learning on classified patches (FDLCP). (a) a phantom image, (b) a zoomed-in region; (c) a class of patches with diagonal geometric direction; (d) non-adaptive 3 level 2D Haar wavelets; (e) atoms of dictionary learning without patch classification; (f) atoms of dictionary learning on patches with diagonal geometric directions in (c), (g) the sparse approximation errors by preserving the largest coefficients. Red lines in (a) and (b) indicate geometric directions of patches. Patches are in size $8 \times 8$.

The sparse coding sub-problem is

$$\mathbf{A}^{(k+1)} = \min_{\mathbf{A}} \left\| \mathbf{X}_\omega - \mathbf{D}_\omega \mathbf{A}^{(k)} \right\|_F^2 + \eta^2 \left\| \mathbf{A}^{(k)} \right\|_0 \quad (6)$$

which is solved by the hard thresholding operation

$$\mathbf{A}^{(k+1)} = H_\eta \left( \mathbf{D}_\omega^H \mathbf{X}_\omega \right) \quad (7)$$

where hard thresholding $H_\eta(\cdot)$ is defined as

$$H_\eta(c) = \begin{cases} c, & |c| \geq \eta \\ 0, & |c| < \eta \end{cases} \quad (8)$$

and $\eta = 0.2$ empirically achieve promising performances for all experiments in this work.

The dictionary updating sub-problem is

$$\min_{\mathbf{D}_\omega} \left\| \mathbf{X}_\omega - \mathbf{D}_\omega \mathbf{A} \right\|_F^2 \quad s.t. \quad \mathbf{D}_\omega^H \mathbf{D}_\omega = \mathbf{I}, \quad (9)$$

that is solved by

$$\mathbf{D}_\omega^{(k+1)} = \mathbf{P} \mathbf{V}^H \quad (10)$$

where $\mathbf{P}$ and $\mathbf{V}$ are orthogonal matrices of the following SVD

$$\mathbf{X}_\omega \mathbf{A}^H = \mathbf{P} \mathbf{\Lambda} \mathbf{V}^H \quad (11)$$

The process of FDLCP is summarized in Algorithm 1. The 2D Haar wavelets, constructed via the tensor product of the corresponding 1D Haar wavelets [38], are used as the initial dictionary $\mathbf{D}_\omega^{(0)}(\omega = \omega_1, \omega_2, \cdots \omega_Q)$. The sparse representation ability of the dictionary is quantified by computing relative $l_2$

**Algorithm 1.** Fast dictionary learning on classified patches

**Initialize**: Set the initial dictionary $\mathbf{D}_\omega^{(0)}(\omega = \omega_1, \omega_2, \cdots \omega_Q)$
**Main**:
1. Estimate geometric directions of patches as (4);
2. Form classes of patches sharing the same direction;
3. For each geometric direction $\omega \in \{\omega_1, \cdots, \omega_Q\}$
   For iterations $k$ = 1, 2, …, $K$
   4. Do the sparse coding as (7);
   5. Run the SVD as (11);
   6. Update the dictionary as (10);
   7. If converge,
       save one class dictionary $\mathbf{D}_\omega$;
     else,
       go to step 4;
8. Output multi-class dictionaries $\mathbf{D}_\omega^{(0)}(\omega = \omega_1, \omega_2, \cdots \omega_Q)$.

norm error (RLNE) [21, 28] of the sparse approximation when a certain percentage of the largest coefficients are preserved. The definition of RLNE is in (21). A lower error implies the approximation is more consistent to the ground-truth image and the dictionary has a better sparsifying ability.

The advantage of classifying patches is illustrated in Fig. 1. It shows that the sparsest representation is obtained using the proposed FDLCP. The trained dictionaries using FDL and FDLCP are adapted to the image thus provide sparser representation than the non-adaptive 2D Haar wavelets (Fig. 1 (d)) and 2D discrete cosine transform (DCT). The trained dictionary using FDL (Fig. 1(e)) represents main directions of all patches but may not sufficiently capture one direction contained in single class of patches due to the orthogonality of dictionary. On the contrary, the trained dictionary (Fig. 1(f)) using FDLCP tends to fit patches with a specific geometrical direction. Therefore, the proposed FDLCP achieves the sparsest representation in Fig. 1(g), where it leads to the fastest decay of approximation error. It is also found that using the proposed FDLCP dictionaries in CS-MRI provide better image reconstruction than that using other pre-defined transforms including Curvelets and Contourlets (See Appendix B), which implies the proposed FDLCP achieves a sparser representation.

*B. Sparse Reconstruction Model with Multi-class Dictionary*

Equipped with the trained dictionaries, we are ready for the undersampled MR image reconstruction.

Let $\mathbf{D}_{\omega_j}^H, \forall j$ is an analysis dictionary with the geometrical direction $\omega$ for the $j^{th}$ image patch, and $\mathbf{R}_j \in \mathbb{R}^{n^2 \times N^2}$ is an operator that extracts the $j^{th}$ image patch $\mathbf{x}_j = \mathbf{R}_j \mathbf{x} \in \mathbb{C}^{n^2}$, $(j=1,2,\ldots,J)$ from the image $\mathbf{x}$. An MR image is reconstructed by solving the following minimization problem:

$$\min_{\mathbf{x}} \sum_{j=1}^{J} \left\| \mathbf{D}_{\omega_j}^H \mathbf{R}_j \mathbf{x} \right\|_1 \quad s.t. \quad \left\| \mathbf{y} - \mathbf{F}_U \mathbf{x} \right\|_2 \leq \varepsilon. \quad (12)$$

The $l_2$ norm term in (12) enforces the fidelity of the reconstruction to the undersampled k-space data. The $l_1$ norm term promotes the patches sparse representation with respect to trained dictionaries. Here we switch the $l_0$ norm in the dictionary training in (5) to the $l_1$ norm in the reconstruction so as to assure solving convex optimization problem. Reconstruction with the non-convex $l_0$ norm penalty can improve the reconstruction as it is discussed in Section IV and Appendix C.

In this work, the overlapping patches are extracted from the image with a shift of one pixel. We assume that they meet the periodic boundary condition, thus satisfying the property:

$$\sum_{j=1}^{J} \mathbf{R}_j^T \mathbf{R}_j = c\mathbf{I}, \quad (13)$$

where $c$ is the overlap factor, meaning that the times of pixels belonging to any patches are the same for all pixels [21]. For a typical patch size 8×8, $c = 64$ is set for the proposed method.

Define a transform $\mathbf{\Psi} = [\mathbf{R}_1^T \mathbf{D}_{\omega_1}, \mathbf{R}_2^T \mathbf{D}_{\omega_2}, \cdots, \mathbf{R}_J^T \mathbf{D}_{\omega_J}]^H$ that satisfies

$$\mathbf{\Psi}^H \mathbf{\Psi} = [\mathbf{R}_1^T \mathbf{D}_{\omega_1}, \mathbf{R}_2^T \mathbf{D}_{\omega_2}, \cdots, \mathbf{R}_J^T \mathbf{D}_{\omega_J}] \begin{bmatrix} \mathbf{D}_{\omega_1}^H \mathbf{R}_1 \\ \mathbf{D}_{\omega_2}^H \mathbf{R}_2 \\ \vdots \\ \mathbf{D}_{\omega_J}^H \mathbf{R}_J \end{bmatrix}$$

$$= \sum_{j=1}^{J} \mathbf{R}_j^T \mathbf{D}_{\omega_j} \mathbf{D}_{\omega_j}^H \mathbf{R}_j = \sum_{j=1}^{J} \mathbf{R}_j^T \mathbf{R}_j = c\mathbf{I}$$

and let $\mathbf{\Phi} = \frac{1}{\sqrt{c}} \mathbf{\Psi}$, one has

$$\mathbf{\Phi}^H \mathbf{\Phi} = \mathbf{I}. \quad (14)$$

This shows that the rows of $\mathbf{\Phi}$ form a tight frame in image space. Therefore, the proposed FDLCP is actually an adaptive tight frame construction method [38]. With (14), MR image reconstruction model in (12) can be rewritten as

$$\min_{\mathbf{x}} \left\| \mathbf{\Phi} \mathbf{x} \right\|_1 \quad s.t. \quad \left\| \mathbf{y} - \mathbf{F}_U \mathbf{x} \right\|_2 \leq \varepsilon. \quad (15)$$

This model means that the target MR image is reconstructed by enforcing its sparsity under a transform embedded with the geometrical directions and trained dictionaries.

How to solve (15) numerically is presented below.

*C. Numerical Algorithm*

To solve the equation (15), we follow the split Bregman for tight frame image restoration [53]. First, an auxiliary variable $\mathbf{\alpha} = \mathbf{\Phi}\mathbf{x}$ is introduced to split the $l_1$ norm and $l_2$ norm terms. Equation (15) is equivalent to

$$\min_{\mathbf{x}, \mathbf{\alpha}} \left\| \mathbf{\alpha} \right\|_1 \quad s.t. \quad \left\| \mathbf{y} - \mathbf{F}_U \mathbf{x} \right\|_2 \leq \varepsilon, \quad \mathbf{\alpha} = \mathbf{\Phi}\mathbf{x} \quad (16)$$

Then we utilize alternating direction method of multipliers (ADMM) [54] to solve (16) according to

$$L_{\lambda, \beta}(\mathbf{x}, \mathbf{\alpha}, \mathbf{d}, \mathbf{h}) = \left\| \mathbf{\alpha} \right\|_1 + \mathbf{h}^T (\mathbf{y} - \mathbf{F}_U \mathbf{x}) + \frac{\lambda}{2} \left\| \mathbf{y} - \mathbf{F}_U \mathbf{x} \right\|_2^2$$
$$+ \mathbf{d}^T (\mathbf{\Phi}\mathbf{x} - \mathbf{\alpha}) + \frac{\beta}{2} \left\| \mathbf{\Phi}\mathbf{x} - \mathbf{\alpha} \right\|_2^2 \quad (17)$$

with an early stopping criteria $\left\| \mathbf{y} - \mathbf{F}_U \mathbf{x} \right\|_2 \leq \varepsilon$. This approach has been previously used in [53, 55]. In practice, we find that

$\varepsilon=10^{-4}$ leads to promising results for all the imaging data used in this paper.

The ADMM technique turns (15) into iteratively solving following sub-problems:

$$\boldsymbol{\alpha}^{(k+1)} = \arg\min_{\boldsymbol{\alpha}} \|\boldsymbol{\alpha}\|_1 + \frac{\beta}{2}\|\boldsymbol{\Phi}\mathbf{x}^{(k)} - \boldsymbol{\alpha} - \mathbf{d}^{(k)}\|_2^2$$

$$\mathbf{x}^{(k+1)} = \arg\min_{\mathbf{x}} \frac{\lambda}{2}\|\mathbf{UFx} - \mathbf{y} - \mathbf{h}^{(k)}\|_2^2 + \frac{\beta}{2}\|\boldsymbol{\Phi}\mathbf{x} - \boldsymbol{\alpha}^{(k+1)} - \mathbf{d}^{(k)}\|_2^2 \quad (18)$$

$$\mathbf{h}^{(k+1)} = \mathbf{h}^{(k)} - \delta_h(\mathbf{UFx}^{(k+1)} - \mathbf{y})$$

$$\mathbf{d}^{(k+1)} = \mathbf{d}^{(k)} - \delta_d(\boldsymbol{\Phi}\mathbf{x}^{(k+1)} - \boldsymbol{\alpha}^{(k+1)})$$

where $\delta_h$ and $\delta_d$ are two constant step sizes and are set as 1. For fixed $\mathbf{x}^{(k)}$, $\mathbf{d}^{(k)}$ and $\mathbf{h}^{(k)}$, $\boldsymbol{\alpha}^{(k+1)}$ is obtained via soft thresholding:

$$\boldsymbol{\alpha}^{(k+1)} = S_{1/\beta}(\boldsymbol{\Phi}\mathbf{x}^{(k)} - \mathbf{d}^{(k)})$$
$$= \max(|\boldsymbol{\Phi}\mathbf{x}^{(k)} - \mathbf{d}^{(k)}| - 1/\beta, 0)\frac{\boldsymbol{\Phi}\mathbf{x}^{(k)} - \mathbf{d}^{(k)}}{|\boldsymbol{\Phi}\mathbf{x}^{(k)} - \mathbf{d}^{(k)}|} \quad (19)$$

For fixed $\boldsymbol{\alpha}^{(k+1)}$, $\mathbf{d}^{(k)}$ and $\mathbf{h}^{(k)}$, $\mathbf{x}^{(k+1)}$ has a close-form solution:

$$\mathbf{x}^{(k+1)} = \mathbf{F}^H(\lambda \mathbf{U}^T\mathbf{U} + \beta\mathbf{I})^{-1}(\lambda \mathbf{U}^T(\mathbf{y} + \mathbf{h}^{(k)}) + \beta\mathbf{F}\boldsymbol{\Phi}^H(\boldsymbol{\alpha}^{(k+1)} + \mathbf{d}^{(k)})) \quad (20)$$

The numerical algorithm is summarized in Algorithm 2.

---

**Algorithm 2** MR image reconstruction with FDLCP.
**Initialize**: Input the undersampled k-space data $\mathbf{y}$, undersampled Fourier encoding matrix $\mathbf{F}_U$, the trained multi-class dictionaries $\{\mathbf{D}_{\omega_j}^H\}$ and the class membership of image patches; Initialize $\mathbf{x} = \mathbf{F}_U^H\mathbf{y}$.
**Main**:
Do
1. Compute the sparse coefficients $\boldsymbol{\alpha}$ by using (19);
2. Update $\mathbf{x}$ by solving normal equation (20);
3. Update multiplier $\mathbf{d}$ and $\mathbf{h}$ by using (18);

Until $\|\mathbf{y} - \mathbf{F}_U\mathbf{x}\|_2 \le \varepsilon$
**Output**: The reconstructed image $\mathbf{x}$.

Note: The stopping condition is checked after the 1st inner iteration.

---

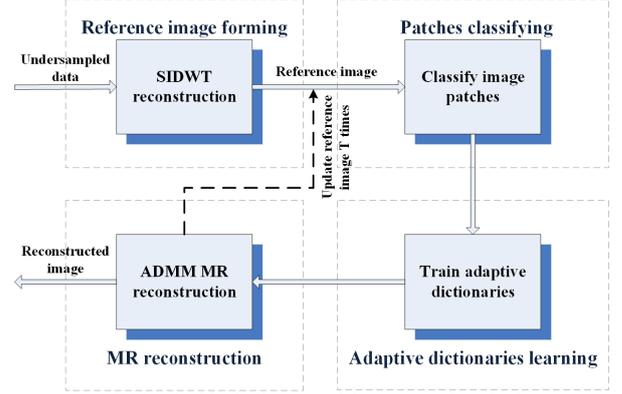

Fig. 2. The block diagram of the proposed method

### D. The Complete Procedure of The Proposed Method

The complete procedure of the proposed method is shown in Fig. 2. It consists of four stages: Reference image forming, patch classification, dictionaries learning and sparse MR image reconstruction. First, a reference image is reconstructed from undersampled k-space data by solving the reconstruction problem in (1) with the shift invariant discrete wavelet (SIDWT) [21, 28] as the sparsifying transform and the ADMM numerical algorithm [54]. Second, geometrical directions are estimated on patches of the reference image and patches sharing the same direction belong to the same class. Third, one dictionary is trained in each single class and multi-class dictionaries are constructed for all classes. Last, image is reconstructed using the multi-class dictionaries.

As the initial reference image usually contains obvious artifacts that may reduce the accuracy of patch classification, the reference may be updated once again for patch classification and dictionary learning, and further improve the reconstruction. Using the SIDWT to obtain the first reference images is not new and has been used in CS-MRI before [21, 28]. Effect of the initial reference is discussed in Section V.

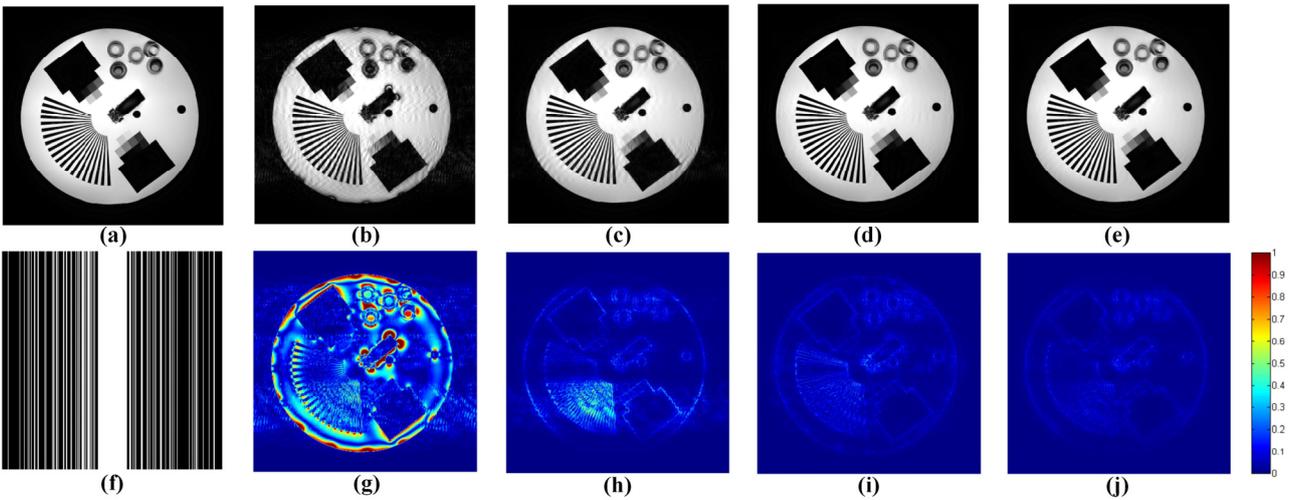

Fig. 3. Reconstructed phantom images and errors using Cartesian sampling pattern when 33% data are sampled. (a) A full sampled phantom image; (b-e) Reconstructed images based on WaTMRI, DLMRI, PBDW and FDLCP, respectively; (f) Cartesian undersampling pattern; (g-j) the reconstruction error magnitudes corresponding to the above reconstructions.

## IV. RESULTS

In this section, image reconstructions on phantom and *in vivo* MR data are carried out to evaluate the performance of the proposed method. Cartesian sampling with random phase encoding [3], 2D random sampling [3, 21, 28] and pseudo radial sampling [16] are adopted here. The proposed FDLCP method is compared with three state-of-the-art CS-MRI methods: WaTMRI [37] which utilizes the wavelet tree sparsity in MR images, DLMRI [20] which is a typical dictionary learning method in CS-MRI, and PBDW [21] which enforces the sparsity using patch-based directional wavelets. We utilize the same zero-filling image $\mathbf{x}^{(0)} = \mathbf{F}_\mathbf{U}^H \mathbf{y}$, which is the original default setting, as the initial numerical solution for all methods. It is worth noting that using SIDWT-based reconstruction to initialize iterative reconstruction leads to faster convergence for these compared methods in practice.

For the WaTMRI, we use the code available at the authors' website [56], and set the total variation parameter $\alpha$=0.001 and the tree sparsity parameter $\beta$=0.04, which work optimally in our experiments. For the PBDW, we utilize the default parameters in our available code [57]. Regarding DLMRI [58], we also set the image patches size as 8×8 ($n$=8) and overlap stride $r$=1. It is worth noting that increasing the overcompleteness of dictionary in DLMRI can significantly improve image reconstruction quality but introduce more computation (See Appendix D). As it is also observed in [20] (See Fig. 13(c) in [20]), the number of dictionary atoms in DLMRI is compromised with the computation. In DLMRI, a square dictionary ($K=n^2=64$) is learnt from 19200 randomly selected patches using 36 iterations. The sparsity level of patches $s$=13 is employed together with an error threshold and the error thresholds for sparse coding varies linearly from 0.046 to 0.032 within 36 iterations. In the following texts, the DLMRI refers to Square DLMRI that using the square dictionary.

In all FDLCP experiments, we use 3-level Daubechies wavelets in SIDWT [28] to obtain the reference image and the times of updating reference image $T$=1, set each image patch size as 8×8 ($n$=8) with maximum patch overlap $c$=64. We pre-define $Q$=71 different geometrical directions for patch classification which is also set in PBDW [21].

Reconstruction performance is quantified by the relative $l_2$ norm error (RLNE) [21, 28] and structure similarity index (SSIM) [59]. The relative $l_2$ norm error (RLNE) [21, 28] is defined as

$$\text{RLNE} = \|\hat{\mathbf{x}} - \mathbf{x}\|_2 / \|\mathbf{x}\|_2 \quad (21)$$

to measure the difference between the reconstructed image $\hat{\mathbf{x}}$ and the fully sampled image $\mathbf{x}$. A lower error implies the reconstructed image is more consistent to the fully sampled image. In our experience, a RLNE that is lower than 0.1 corresponds to an acceptable reconstruction quality. The structure similarity index (SSIM) [59] is defined as

$$\text{SSIM}(\mathbf{x}, \hat{\mathbf{x}}) = \frac{(2\mu_\mathbf{x}\mu_{\hat{\mathbf{x}}} + c_1)(2\sigma_{\mathbf{x}\hat{\mathbf{x}}} + c_2)}{(\mu_\mathbf{x}^2 + \mu_{\hat{\mathbf{x}}}^2 + c_1)(\sigma_\mathbf{x}^2 + \sigma_{\hat{\mathbf{x}}}^2 + c_2)} \quad (22)$$

where $\mu_\mathbf{x}$, $\mu_{\hat{\mathbf{x}}}$, $\sigma_\mathbf{x}$, $\sigma_{\hat{\mathbf{x}}}$ and $\sigma_{\mathbf{x}\hat{\mathbf{x}}}$ are the means, standard deviations and covariance for the images $\mathbf{x}$ and $\hat{\mathbf{x}}$, $c_1 = (k_1 L)^2$, $c_2 = (k_2 L)^2$ are two constant variables to avoid instability when the denominator $\mu_\mathbf{x}^2 + \mu_{\hat{\mathbf{x}}}^2$ or $\sigma_\mathbf{x}^2 + \sigma_{\hat{\mathbf{x}}}^2$ close to zero, $k_1$=0.01, $k_2$=0.03 are two small constant, $L$ is the dynamic range of the pixel. A higher SSIM indicates two images have more structural similarity, which means the stronger detail preservation in reconstruction [18].

### A. Experiments on Phantom Data

Fig. 3 shows reconstruction on a phantom data which contains a lot of geometrical directions. The fully sampled phantom (Fig. 3(a)) is acquired from 3T Siemens MRI scanner using a turbo spin echo sequence (matrix size = 384×384, TR/TE=2000/9.7ms, field of view = 230×187mm$^2$, slice thickness = 5.0mm).

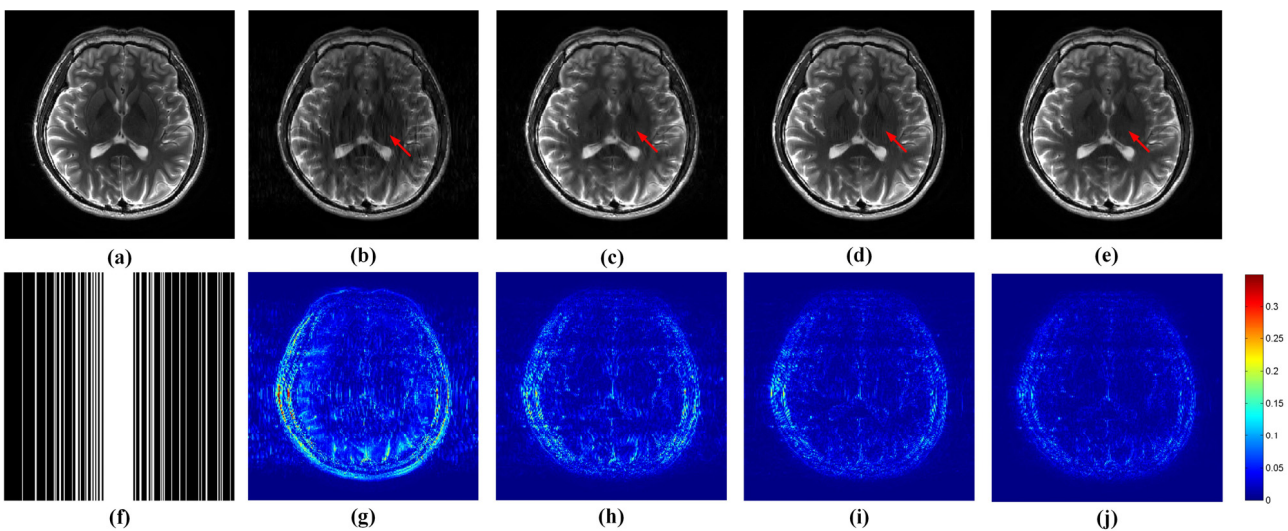

Fig. 4. Reconstructed brain images and errors using Cartesian sampling pattern with sampling rate 0.32. (a) A full sampled brain image; (b-e) Reconstructed images using WaTMRI, DLMRI, PBDW and FDLCP, respectively; (f) Cartesian undersampling pattern; (g-j) the reconstruction error magnitudes corresponding to the above reconstructions.

WaTMRI introduces obvious artifacts whereas DLMRI causes ringing around the edges. PBDW reconstructs images much better but produces artifacts in the smooth region in Fig. 3(d) and loses some edges in Fig. 3(i). The proposed FDLCP reconstructs the image best in Fig. 3(e) and leads to minimal loss of image features in Fig. 3(j). The quality metrics listed in TABLE I implies that FDLCP achieves the lowest RLNE and highest SSIM among all methods.

TABLE I
RLNE/SSIM RESULTS FOR RECONSTRUCTED PHANTOM AND BRAIN

| Images | WaTMRI | DLMRI | PBDW | FDLCP |
|---|---|---|---|---|
| Phantom | 0.1542 /0.7164 | 0.0885 /0.8693 | 0.0461 /0.9681 | **0.0315 /0.9846** |
| Brain | 0.1607 /0.8243 | 0.1414 /0.8650 | 0.1145 /0.9468 | **0.0959 /0.9630** |

Note: MR images in Fig. 3(a) and Fig. 4(a) are used for experiments.

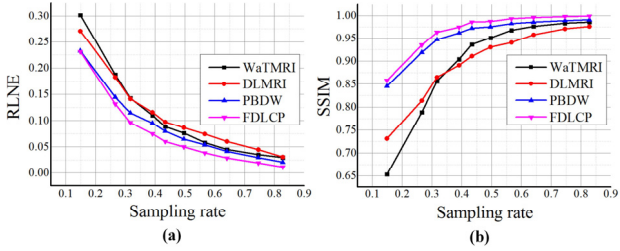

Fig. 5. Brain image reconstruction qualities versus different sampling rates. (a) and (b) are RLNE and SSIM versus different sampling rates. Note: T2-weighted image in Fig. 4(a) is used for experiments.

### B. Experiments on Brain Imaging Data

The T2-weighted and T1-weighted brain imaging data are obtained from different scanners. T2-weighted brain images, Fig. 4(a) and Fig. 7(d), are two slices acquired from a healthy volunteer at a 3T Siemens Trio Tim MRI scanners using the T2-weighted turo spin echo sequence (matrix size = 256×256, TR/TE=6100/99ms, field of view = 220×220mm$^2$, slice thickness = 3.0mm). Fig. 6(a) is another T2-weighted image measured from a healthy volunteer at another 3T Siemens scanner using a turbo spin echo sequence (matrix size = 384×324, TR/TE = 5000/97ms, field of view = 230×187mm$^2$, slice thickness = 5.0mm). T1-weighted brain images, Fig. 7(e) and (f), are two slices obtained from a healthy volunteer at 1.5T Philips MRI scanner with fast-field-echo sequences (matrix size = 256×256, TR/TE = 1700/390ms, field of view = 230×230mm$^2$, slice thickness = 5mm).

The reconstruction errors in Fig. 4 show that FDLCP has lowest errors near edges and the fewest aliasing artifacts in the smooth region. Visual inspection is consistent to the two reconstruction metrics. The RLNEs and SSIMs in TABLE I point out that FDLCP leads to the lowest reconstruction error and highest reconstruction structure similarity among four reconstruction methods.

At different sampling rates, when the same Cartesian sampling schemes are used, consistent reductions on the RLNE and improvement on the SSIM are observed in Fig. 5. The proposed FDLCP outperforms other state-of-the-art methods under this sampling pattern.

We also test the performance of FDLCP with other sampling patterns. Pseudo radial sampling is employed in Fig. 6. The error image of FDLCP is less structured, which indicates that FDLCP preserves the image features better than other methods. Besides, the superior RLNE and SSIM metrics, shown in Fig. 6(g) and (h), also implies the advantage of FDLCP. Another sampling patterns on more brain images are tested in Fig. 7. The RLNE and SSIM metrics are listed in TABLE II, implying that FDLCP always performs better than the compared methods.

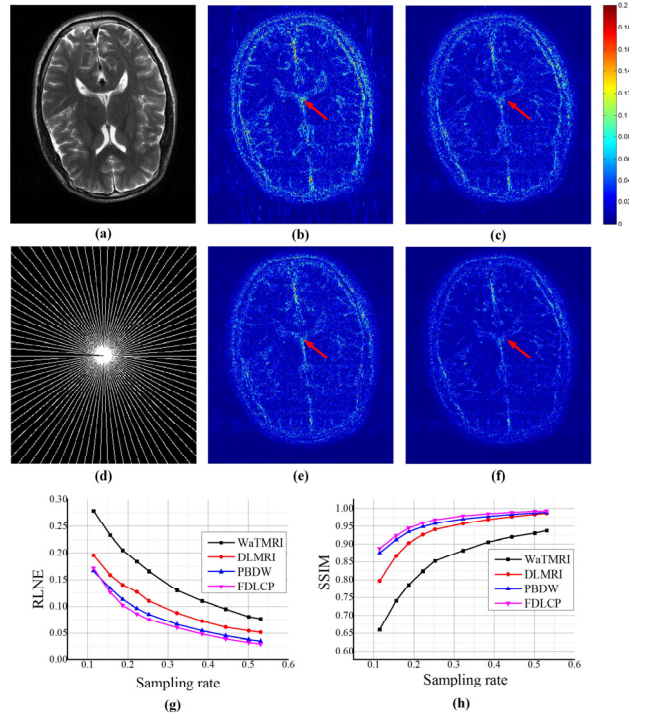

Fig. 6. Reconstruction error using radial sampling pattern. (a) A fully sampled brain image; (b, c, e, f) Reconstruction errors for WaTMRI, DLMRI, PBDW and FDLCP, respectively; (d) Pseudo radial sampling pattern with sampling rate 0.18; (g, h) RLNEs and SSIMs versus different sampling rates, respectively.

TABLE II
RLNE/SSIM RESULTS FOR BRAINS IN FIG. 7 USING THREE SAMPLINGS

| Images | Mask | WaTMRI | DLMRI | PBDW | FDLCP |
|---|---|---|---|---|---|
| Fig. 7(d) | Fig. 7(a) | 0.1993 /0.7192 | 0.1878 /0.7561 | 0.1354 /0.8977 | **0.1295 /0.9174** |
| | Fig. 7(b) | 0.1417 /0.8865 | 0.1499 /0.8581 | 0.1321 /0.9331 | **0.1201 /0.9526** |
| | Fig. 7(c) | 0.1489 /0.8865 | 0.1415 /0.9031 | 0.1130 /0.9343 | **0.1080 /0.9465** |
| Fig. 7(e) | Fig. 7(a) | 0.2259 /0.6958 | 0.1821 /0.8160 | 0.1570 /0.8699 | **0.1402 /0.8937** |
| | Fig. 7(b) | 0.0954 /0.9519 | 0.1015 /0.8732 | 0.0857 /0.9549 | **0.0818 /0.9673** |
| | Fig. 7(c) | 0.0937 /0.9472 | 0.1011 /0.9281 | 0.0782 /0.9614 | **0.0722 /0.9701** |
| Fig. 7(f) | Fig. 7(a) | 0.2191 /0.7129 | 0.1643 /0.8236 | 0.1252 /0.8799 | **0.1125 /0.9216** |
| | Fig. 7(b) | 0.1041 /0.9400 | 0.1217 /0.8250 | 0.0990 /0.9377 | **0.0971 /0.9391** |
| | Fig. 7(c) | 0.1047 /0.9294 | 0.1226 /0.8902 | 0.0890 /0.9405 | **0.0842 /0.9553** |

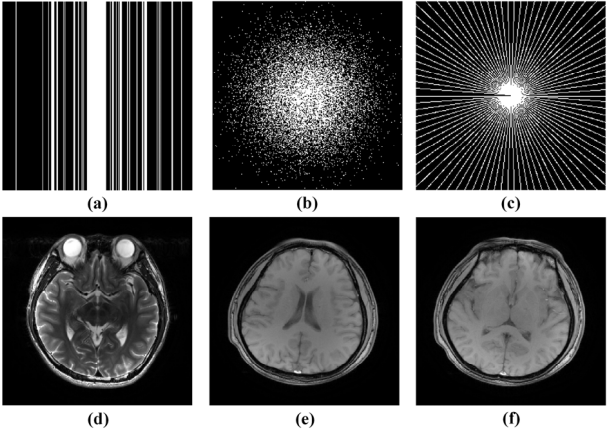

Fig. 7. Three sampling patterns and more brain images. (a) The Cartesian sampling pattern of sampling rate 0.20; (b) The 2D random sampling pattern of sampling rate 0.16; (c) The pseudo radial sampling pattern of sampling rate 0.18; (d) A T2-weighted brain image; (e-f) Two different slices T1-weighted brain images.

## C. Computation Time

All the simulation runs on a 64-bit Window 7 operating system with an Intel Core i5 CPU at 3.30GHz and 12GB RAM. The computation time is obtained by averaging the time of repeated 10 tests. Patches classifying and dictionaries learning are performed twice in the proposed FDLCP.

The running time is listed in Fig. 8. It shows that the proposed FDLCP runs much faster than DLMRI and PBDW but slower than WaTMRI. Compared with WaTMRI, the additional computational cost of FDLCP is acceptable considering its improvement on image reconstruction. Fig. 8 shows the proportions of computation time spent in each stage of FDLCP, which illustrates that reference image forming, patch classification and dictionaries learning are fast and only account for 15 percents of the total time (TABLE III). Note that in FDLCP, the patch classification is optimized using MEX/C code, whereas the dictionary learning and image reconstruction are implemented with Matlab. In PBDW [56], the adaptive sparse representation training and the forward/backward transform are written in MEX/C code, whereas other image reconstruction steps are implemented with Matlab. Both WaTMRI [56] and DLMRI [58] are implemented with Matlab and their runtime can be substantially reduced with MEX/C implementations.

## D. Discussion on Parameter Settings

In this section, we analyze the effect of parameter settings in FDLCP. The brain image in Fig. 4(a) and the undersampling pattern in Fig. 4(f) are adopted in the experiment. The parameter discussed are the patch size ($n \times n$), the number of the geometrical directions ($Q$), the reference image and the times of updating reference image ($T$). Typical settings are $n \times n = 8 \times 8$, $Q=71$, $T=1$ and the reference image is obtained by SIDWT. When one parameter is analyzed, other parameters are set as the typical values.

The optimal patch size is $8 \times 8$. The effect of patch size is shown in Fig. 9(a). When the patch size is increased from $2 \times 2$ to $8 \times 8$, both RLNE and SSIM are improved. Because a larger patch contains more discrete pixels, more possible geometric directions can be estimated. This allows more accurate patch classification and better sparsity for patches within each class. However, when patch size is too large, e.g. $16 \times 16$, some patches may contains multiple directions of edges, but only one dominant geometric direction is estimated for each class of patches. In this case, the trained dictionary will be hard to sufficiently capture geometrical directions of these patches and the sparsity is reduced, resulting in degraded image reconstruction.

The reconstruction performance is not sensitive to the number of geometrical directions $Q$. RLNE and SSIM are slightly changed when the number of geometrical directions varies in $Q \in [8, 71]$ as shown in Fig. 9(b). To maximally explore the geometric directions, $Q=71$ is set for the patch size $8 \times 8$.

The optimal times of updating reference image $T$ is 1. Since the FDLCP reconstructs the image much better than SIDWT-based reference image, more accurate patch classification and sparser representation will be achieved when the times of updating reference image increases. As shown in Fig. 10(g, h), RLNE and SSIM are significantly improved as $T$ increases from 0 to 1. When $T>1$, the improvement is marginal but updating the reference image costs more computation. Therefore, we update reference images using one time of SIDWT and FDLCP reconstruction to tradeoff image quality and computation time.

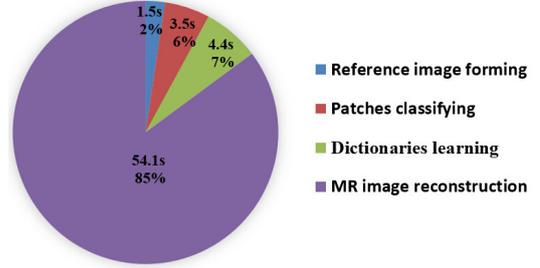

TABLE III
COMPUTATION TIME OF DIFFERENT METHODS

| Reconstruction methods | WaTMRI | DLMRI | PBDW | FDLCP |
|---|---|---|---|---|
| Computation time (s) | **1.4** | 2123.3 | 425.7 | 63.5 |

Fig. 8. Computation time of each stage in the proposed FDLCP.

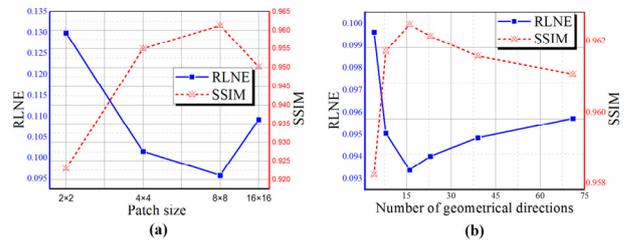

Fig. 9. The performance of FDLCP with various parameter setting. (a) and (b) are RLNE and SSIM versus the patch size and the number of geometrical directions.

## E. Effect of Initial Reference Image

The proposed method is not sensitive to initial reference images as shown in Fig. 10(g, h). SIDWT-based reference ($T=0$) leads to higher quality image ($T=1$) (Fig. 10(b)) than that zero-filling reference (Fig. 10(e)). When the times of updating reference image is 2, FDLCP-based reconstruction ($T$=3) using two different initial reference images ($T$=2) are comparable (Fig 10(c, f)) and evaluation metrics are nearly the same ($T$=3 in Fig. 10(g, h)). This means using SIDWT-based reference will require 1 time of SIDWT and FDLCP reconstruction but zero-filling-based reference needs 2 times of FDLCP reconstruction. We use the SIDWT-based reference image for a good start-up.

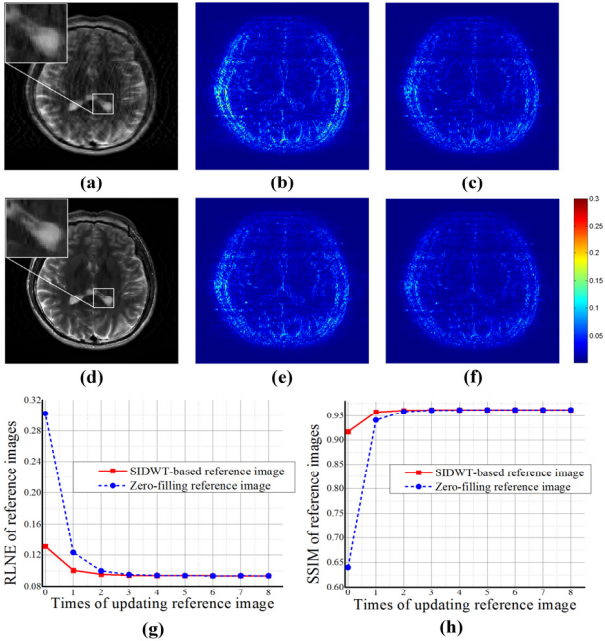

Fig. 10. Reconstructed images using different references and different times of updating reference image. (a) is the initial reference image obtained using zero-filling ($T$=0), (b, c) are the reconstruction error magnitudes of FDLCP-based reconstruction using the zero-filling reference image when $T$=0 and 2, respectively; (d) is the initial reference image obtained using SIDWT ($T$=0), (e, f) are the reconstruction error magnitudes of FDLCP-based reconstruction using the SIDWT-based reference image when $T$=0 and 2, respectively; (g, h) are the evaluation metrics, RLNE and SSIM, versus the times of updating reference image. Note: Along the horizontal axis in (g) and (h), "0" corresponds to quality of initial reference and other numbers "1"-"8" correspond to quality of FDLCP-based reconstructions using the reference images corresponding to "0"-"7", respectively.

*F. Comparison with Other State-of-the-Art Methods*

We carry out the comparisons between FDLCP and more recent MR reconstruction methods: The PBDWS [18] that enhances the PBDW [21] by extending directional wavelet into the shift-invariant discrete wavelet domain; The BPFA [60] which uses the beta process to learn a nonparametric dictionary; The PANO [28] that forms a general patch-based nonlocal operator to sparsely represent the similar patches; The NLS [61] that introduces a fast iterative non-local shrinkage algorithm. Both FDLCP and PANO solve convex problems while PBDWS, BPFA and NLS solve non-convex problems. Since non-convex methods are observed to improve the image reconstruction [16, 18], we also implement a non-convex version of the proposed FDLCP for a fair comparison by replacing the $l_1$ norm with $l_0$ norm in the reconstruction model in (15) (See Appendix C).

We use the built-in parameter settings in PBDWS, BPFA, and PANO implementations since the same brain imaging data are used in these methods. For the NLS, we use the $l_{0.5}$ non-local shrinkage penalty and set the regularization parameter $\lambda=10^{-4}$, the number of inner iterations and outer iterations are 10 and 35, respectively.

TABLE IV indicates that the $l_1$ norm FDLCP achieves lower reconstruction error than PANO whereas $l_0$ norm FDLCP obtains better reconstruction than PBDWS, BPFA and NLS. BPFA outperforms PBDWS and NLS for T2-weighted brain images in Fig. 4(a) and Fig. 7(d), whereas NLS beats BPFA for the T1-weighted brain image in Fig. 7(f). TABLE V lists the computation time of these methods. The fastest method is NLS and the slowest is BPFA. The FDLCP is relatively faster than PANO and PBDWS. These metrics indicate that FDLCP leads to competitive performance. Note that in PANO [62] and PBDWS [63] methods, the adaptive sparse representation training and the forward/backward transforms are written in MEX/C code, whereas other image reconstruction steps are implemented with Matlab. Both NLS and BPFA are implemented with Matlab and their computation time can be substantially reduced with MEX/C implementations

TABLE IV
RLNE/SSIM RESULTS FOR RECONSTRUCTED MR IMAGE USING OTHER STATE-OF-THE-ART METHODS

| Image | Convex | | Non-convex | | | |
|---|---|---|---|---|---|---|
| | PANO | FDLCP $l_1$ norm | PBDWS | BPFA | NLS | FDLCP $l_0$ norm |
| Fig. 4 (a) | 0.0978 /0.9404 | **0.0935** /**0.9626** | 0.0937 /0.9493 | 0.0893 /0.9600 | 0.1190 /0.6875 | **0.0778** /**0.9707** |
| Fig. 7 (d) | 0.1179 /0.9322 | **0.0916** /**0.9585** | 0.1021 /0.9383 | 0.0956 /0.9531 | 0.1225 /0.6881 | **0.0875** /**0.9592** |
| Fig. 7 (f) | 0.0967 /0.9300 | **0.0769** /**0.9601** | 0.0830 /0.9454 | 0.0909 /0.9227 | 0.0742 /0.9562 | **0.0722** /**0.9580** |

Note: The sampling mask shown in Fig. 4(f) is adopted in experiment.

TABLE V
COMPUTATION TIME FOR MR IMAGE RECONSTRUCTIONS USING OTHER STATE-OF-THE-ART METHODS

| Method | Convex | | Non-convex | | | |
|---|---|---|---|---|---|---|
| | PANO | FDLCP $l_1$ norm | PBDWS | BPFA | NLS | FDLCP $l_0$ norm |
| Computation time (s) | 130.2 | **56.4** | 142.4 | 3182.9 | **41.4** | 60.2 |

## V. CONCLUSION

A new MR image reconstruction method based on fast multi-class dictionaries learning is proposed. Image patches are classified according to their geometrical directions, and orthogonal dictionaries are trained within each class. The alternating direction method of multipliers is adopted to reconstruct the image efficiently. Results on phantom and brain imaging data demonstrate the superior performance of the proposed method in suppressing artifacts and preserving image edges. The proposed method outperforms the compared state-of-the-art MR image reconstruction methods and its computation is much faster than typical K-SVD dictionary learning methods. How to classify patches with multiple features, not only geometrical directions, to provide sparser image representation will be further developed. Besides, since the trained dictionaries form a tight frame, a recent projected fast iterative soft-thresholding algorithm [64] can be utilized for fast and stable image reconstruction.

ACKNOWLEDGMENTS

The authors would like thank Bingwen Zheng, Feng Huang and Xi Peng for providing data used in this paper. The authors also thank the following scholars for sharing codes for comparisions: Yue Huang and Xinghao Ding for BPFA, Yasir Q. Mohsin and Mathews Jacob for NLS, Junzhou Huang for WaTMRI, Saiprasad Ravishankar, Bihan Wen and Yoram Bresler for DLMRI. The authors are grateful to the editors and reviewers for their constructive comments.

# Appendix

## APPENDIX A
### THE DETAILED DESCRIPTION OF OPERATOR $\mathbf{G}_\omega$

The $\mathbf{G}_\omega$ is an operator that re-arranges pixels along a candidate geometrical directions [1]. All the candidate geometric directions are pre-defined in a patch and are uniformly partitioned in the interval $[0, 2\pi]$. The candidate directions are marked with the white lines shown in Fig. 1(a) and only 14 directions are presented as an illustrative example. The angle between the direction lines (red color in Fig. 1(b)) and the horizontal direction (dashed line in Fig. 1(b)) stands for these directions. Let the candidate directions be $\{\theta_1, \theta_2, \cdots, \theta_Q\}$ and a specified direction be $\theta_d \in \{\theta_1, \theta_2, \cdots, \theta_Q\}$, there is an associated direction line $L_{\theta_d}$ (red color in Fig. 1(b)) and its orthogonal line $L_{\theta_d}^\perp$ (blue color in Fig. 1(b)). Each patch pixel $x(r_X, r_Y)$ located at $(r_X, r_Y)$ is orthogonally projected onto the line $L_{\theta_d}^\perp$ to get a new point $x(r_{\theta_d}^\perp)$, and pixels are reordered by the projected distance along the line $L_{\theta_d}^\perp$. Finally, 64 pixels are used to produce a 1D column vector according to the order marked on each pixel in Fig. 1(c).

By rotating the central line in a 8×8 patch, 71 is the maximal number that determines discrete grids to cover the pixels. This setting allows maximally explore the geometric directions in a patch. As the default parameter setting in PBDW [1], 71 directions are pre-defined for 8×8 patches. We choose 71 directions in FDLCP in order to have a fair comparison to PBDW.

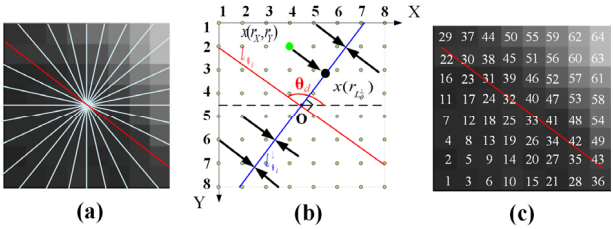

Fig. 1. Illustration of reordering pixels. (a) All candidate directions in a patch; (b) Projecting a pixel to the axis that is orthogonal to a given direction line $L_{\theta_d}$ associated with an angle $\theta_d$; (c) indexes of reordering pixels into 1D vector.

## APPENDIX B
### IMAGE RECONSTRUCTIONS USING DIFFERENT SPARSIFYING DICTIONARIES/TRANSFORMS

We compare our proposed reconstruction method with that using Curvelets [2] or Contourlets [3, 4] as the sparsifying transform in CS-MRI. We directly utilize these transforms to reconstruct MR image. A better reconstructed image implies the dictionary/transform achieves the sparser representation.

The reconstruction model is

$$\min_{\mathbf{x}} \|\mathbf{\Psi x}\|_1 \quad s.t. \quad \|\mathbf{y} - \mathbf{F}_U \mathbf{x}\|_2 \le \varepsilon \quad (A1)$$

where $\mathbf{\Psi}$ is the Curvelets or Contourlets transform. We use the public implementations of these two transforms [4, 5] shared by the respective authors. In the implementation, ADMM [6, 7] is adopted as the numerical algorithm to solve (A1). For Contourlets, we set $2^5$, $2^4$, $2^4$, $2^3$ directional sub-bands from coarse to fine scales, and employ the quincunx-type filter named pkva [8] and no downsampling of the low-pass sub-band at the first level decomposition. For Curvelets, we use wrapping-based fast discrete curvelets transform [5] with 5 decomposition levels and 16 angles at the 2$^{nd}$ coarsest level. Parameters in the reconstruction are tuned to obtain the optimal performance of each transform. Reconstructed images in Fig. 2 show that the proposed FDLCP achieves better image quality than Curvelets and Contourlets.

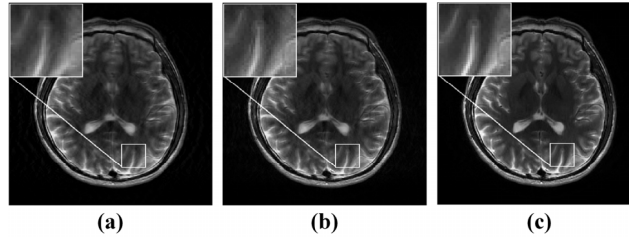

Fig. 2. Reconstructed images using different sparsifying transforms. (a-c) are reconstructed images using Curvelets, Contourlets, and the proposed FDLCP. The reconstruction errors, RLNE, of (a-c) are 0.1634, 0.1589 and 0.0935, respectively. The preserved structure similarity, SSIM, of (a-c) are 0.8407, 0.8572 and 0.9626, respectively. FDLCP achieves best results.

## APPENDIX C
### THE FDLCP USING $L_0$ NORM PENALTY IN RECONSTRUCTION

The reconstruction model of FDLCP using the $l_0$ norm is

$$\min_{\mathbf{x}} \|\mathbf{\Phi x}\|_0 \quad s.t. \quad \|\mathbf{y} - \mathbf{F}_U \mathbf{x}\|_2 \le \varepsilon. \quad (A2)$$

We also use the ADMM [6, 7] to solve the model. The whole process is the same as the Algorithm 2 except the thresholding. The sparse coefficients is obtained by hard thresholding and the threshold is $\sqrt{2/\beta}$ instead. The solution can be expressed as following

$$\mathbf{\alpha}^{(n+1)} = H_{\sqrt{2/\beta}}\left(\mathbf{\Phi x}^{(n+1)} - \mathbf{d}^{(n)}\right) \quad (A3)$$

The $l_0$ norm penalty improves image quality (Fig. 3). The reconstruction error has been reduced by 22%. Although it is hard to prove the convergence theoretically, the curves in Fig. 4 empirically show that RLNE and the objective function in (A2) decreases and gradually stabilizes as the iteration times increase, although there is a small oscillation at the beginning.

The code of the FDLCP with both $l_1$ and $l_0$ norm minimization will be released at the authors' website [9].

APPENDIX D

EFFECT OF THE NUMBER OF DICTIONARY ATOMS IN DLMRI

In this appendix, as the reviewer requested, we add a comparison when the total number of dictionary atoms in DLMRI [10] is set the same as that in the proposed FDLCP, and analyze the effect of the number of dictionary atoms in DLMRI.

For the brain image shown in Fig. 3(a), there are actually 58 different geometrical directions estimated from all the brain image patches, other 13 directions are not found although 71 geometrical directions are typically predefined for a 8×8 patch in FDLCP. Therefore, the total number of atoms in FDLCP and DLMRI is 58×64=3712. The reconstructed images shown in Fig. 5 indicate that FDLCP preserves image edges better than DLMRI and achieve both lower RLNE and higher SSIM. Besides, FDLCP runs much faster (approximately 60s) than DLMRI.

Fig. 5(e-f) show that increasing the number of dictionary atoms in DLMRI can improve the reconstruction but also introduce more computations. When the number of dictionary atoms increases from 1024 to 4096, the improvements on RLNE and SSIM are marginal but the computation time is about 3.5 times. Taking the computation time into account, the number of dictionary atoms is set to 64. The original authors of DLMRI typically set the number of atoms being equal to the number of pixels in a patch [10], which leads to promising results both in their and our experiments.

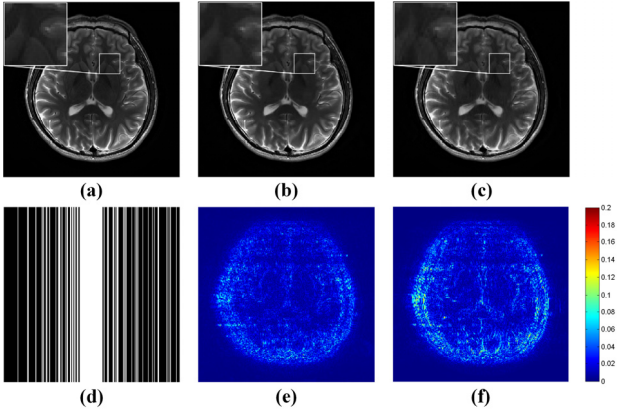

Fig. 3. A comparison of reconstruction images using $l_0$ and $l_1$ norm penalty in FDLCP. (a) A full sampled brain image; (b-c) Reconstructed images using $l_0$ and $l_1$ norm penalties, respectively; (d) Cartesian undersampling pattern with 32% data; (e-f) the reconstruction error magnitudes corresponding to (b-c), respectively; RLNE of (b-c) are 0.0741 and 0.0935, SSIM of (b-c) are 0.9707, 0.9626. Note: The parameters of FDLCP are the patch size 8×8, the pre-defined 71 different geometric directions for patch classification, the regularization parameter $\lambda=10^3$, and the times of updating reference image $T=1$.

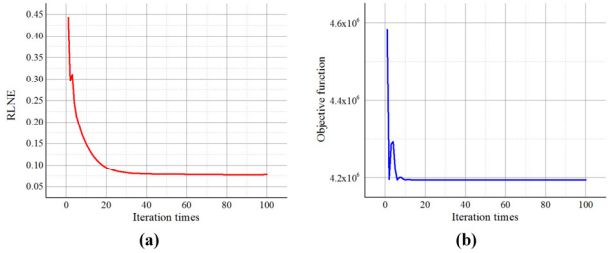

Fig. 4. Empirical convergence of the $l_0$ norm minimization problem. (a) The RLNE between the reconstructed image and the ground truth image versus the iteration time; (b) The values of the objective function in (A2) versus the iteration time.

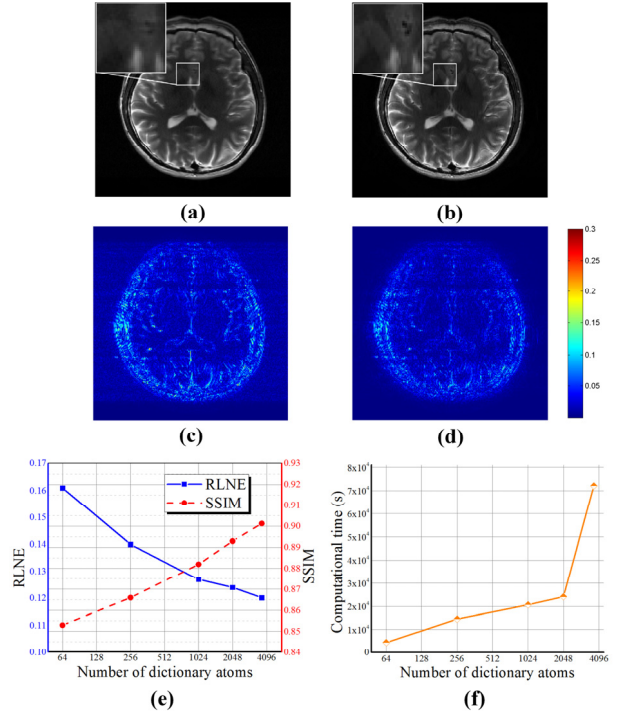

Fig. 5. The effect of the number of dictionary atoms in DLMRI. (a-b) are the reconstructed images using DLMRI and FDLCP when the total number of atoms is set as 3712; (c-d) are reconstruction error magnitudes corresponding to (a-b), respectively; RLNE of (a-b) are 0.1202 and 0.0935, SSIM of (a-b) are 0.9013, 0.9626; (e) RLNE and SSIM versus the number of dictionary atoms with comparison to FDLCP; (f) Computation time versus the number of dictionary atoms.